\documentclass[journal]{vgtc}         % review (journal style)
\ifpdf%                                % if we use pdflatex
  \pdfoutput=1\relax                   % create PDFs from pdfLaTeX
  \usepackage{graphicx}                % allow us to embed graphics files
  \DeclareGraphicsExtensions{.pdf,.png,.jpg,.jpeg} % for pdflatex we expect .pdf, .png, or .jpg files
\else%                                 % else we use pure latex
  \usepackage{graphicx}                % allow us to embed graphics files
  \DeclareGraphicsExtensions{.eps}     % for pure latex we expect eps files
\fi%

\graphicspath{{figures/}{pictures/}{images/}{./}} % where to search for the images

\usepackage{microtype}                 % use micro-typography (slightly more compact, better to read)
\PassOptionsToPackage{warn}{textcomp}  % to address font issues with \textrightarrow
\usepackage{textcomp}                  % use better special symbols
\usepackage{amsmath, bm}

\usepackage{mathptmx}                  % use matching math font
\usepackage{times}                     % we use Times as the main font
\usepackage{cite}                      % needed to automatically sort the references
\usepackage{tabu}                      % only used for the table example
\usepackage{booktabs}                  % only used for the table example
\usepackage{xcolor}
\usepackage{comment}
\usepackage{xspace}
\usepackage{amsmath}
\usepackage{algorithm}
\usepackage[noend]{algpseudocode}
\usepackage{amssymb}
\usepackage{stmaryrd}
\usepackage{trimclip}
\usepackage{multirow}
\usepackage{balance}

\makeatletter
\def\BState{\State\hskip-\ALG@thistlm}
\makeatother

\newcommand{\q}[1]{``#1''}
\newcommand{\x}[1]{\textcolor{black}{#1}}
\newcommand{\rrv}[1]{\textcolor{black}{#1}}

\newcommand{\name}{SeqCausal\xspace}

\newcommand*{\img}[1]{%
    \raisebox{-.3\baselineskip}{%
        \includegraphics[
        height=\baselineskip,
        keepaspectratio,
        ]{#1}%
    }%
}

\makeatletter
\DeclareRobustCommand{\shortto}{%
  \mathrel{\mathpalette\short@to\relax}%
}

\newcommand{\short@to}[2]{%
  \mkern2mu
  \clipbox{{.5\width} 0 0 0}{$\m@th#1\vphantom{+}{\shortrightarrow}$}%
  }
\makeatother

\usepackage{enumitem}

\onlineid{1152}

\vgtccategory{Research}
\vgtcpapertype{Application/Design Study}

\title{Visual Causality Analysis of Event Sequence Data}

\author{Zhuochen Jin, Shunan Guo, Nan Chen, Daniel Weiskopf, David Gotz, Nan Cao}
\authorfooter{
\item
 Zhuochen Jin, Shunan Guo, Nan Chen and Nan Cao are with iDV$^{x}$ Lab at Tongji University. E-mail: \{zcjin.idvx, g.shunan,
 christy05.chen, nan.cao\}@gmail.com. Nan Cao is the corresponding author. 
\item
 Daniel Weiskopf is with University of Stuttgart. E-mail:  weiskopf@visus.uni-stuttgart.de.
\item
 David Gotz is with University of North Carolina at Chapel Hill. E-mail: gotz@unc.edu.
}

\shortauthortitle{Biv \MakeLowercase{\textit{et al.}}: Global Illumination for Fun and Profit}
\abstract{
Causality is crucial to understanding the mechanisms behind complex systems and making decisions that lead to intended outcomes. Event sequence data is widely collected from many real-world processes, such as electronic health records, web clickstreams, and financial transactions, which transmit a great deal of information reflecting the causal relations among event types. Unfortunately, recovering causalities from observational event sequences is challenging, as the heterogeneous and high-dimensional event variables are often connected to rather complex underlying event excitation mechanisms that are hard to infer from limited observations. Many existing automated causal analysis techniques suffer from poor explainability and fail to \rrv{include} an adequate amount of human knowledge. In this paper, we introduce a visual analytics method for recovering causalities in event sequence data. We extend the Granger causality analysis algorithm on Hawkes processes to incorporate user feedback into causal model refinement. The visualization system includes an interactive causal analysis framework that supports bottom-up causal exploration, iterative causal verification and refinement, and causal comparison through a set of novel visualizations and interactions. We report two forms of evaluation: \rrv{a} quantitative evaluation of the model improvements \rrv{resulting from} the user-feedback mechanism, and \rrv{a} qualitative evaluation through \x{case studies \rrv{in} different application domains} to demonstrate the usefulness of the system.

}


\CCScatlist{ % not used in journal version
 \CCScat{K.6.1}{Management of Computing and Information Systems}%
{Project and People Management}{Life Cycle};
 \CCScat{K.7.m}{The Computing Profession}{Miscellaneous}{Ethics}
}

\teaser{
  \centering
  \includegraphics[width=\linewidth]{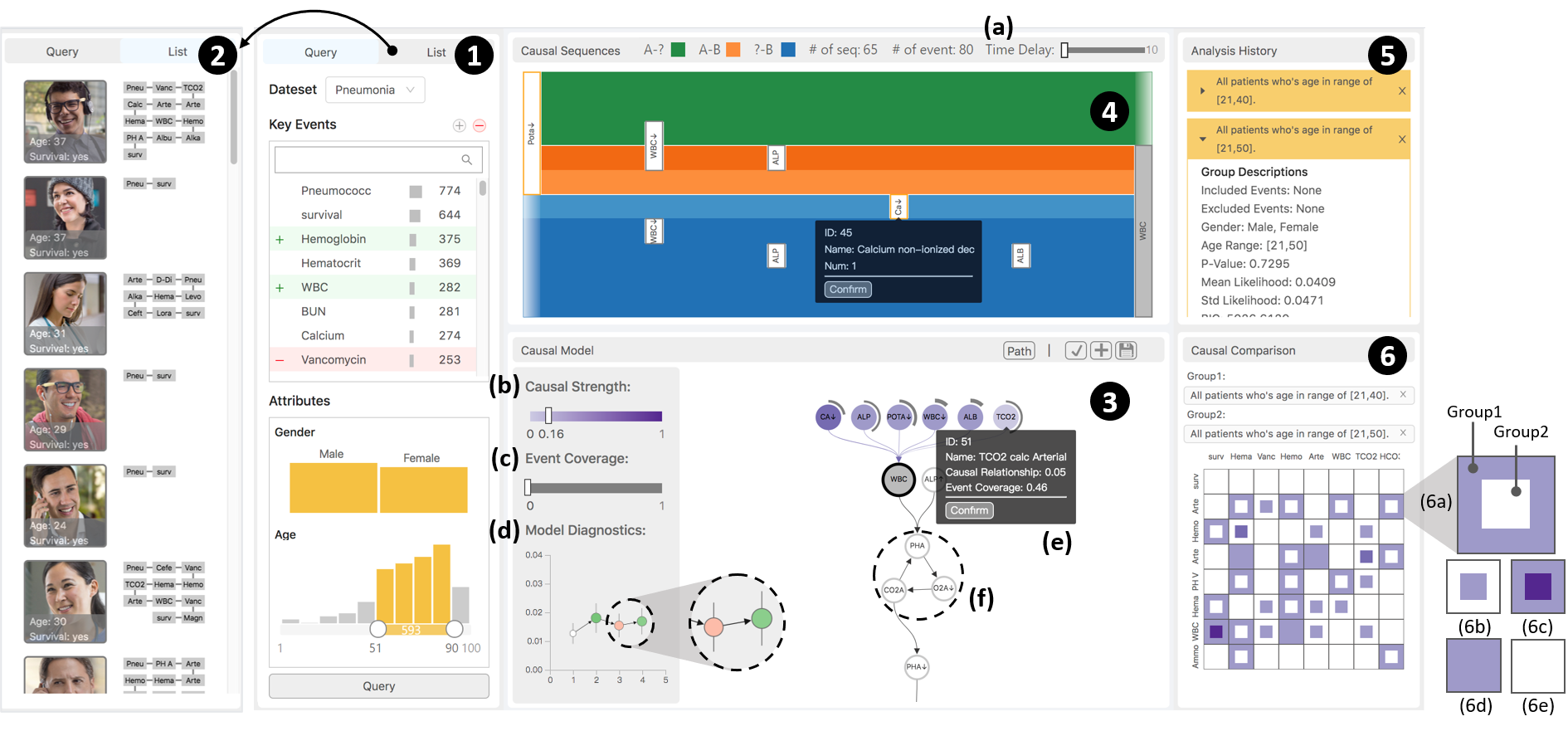}
  \vspace{-8.3mm}
    \caption{An overview of the \name interface. The \textit{query view} (1) provides a set of filters for the user to select sequences for analysis. The \textit{sequence list view} (2) displays individual records retrieved from the query. The \textit{causal model view} (3) displays the causal relations of events calculated from the back-end causality analysis model. Users can modify the graph, \rrv{for example, confirm or delete} a causal link, by examining causal relations from the \textit{causal sequence view} (4), which summarizes causal patterns in raw event sequences. The \textit{analysis history view} (5) stores causalities of different queried subsets, from which users can select any two items to compare their causal relations in the \textit{causal comparison view} (6).}
    \label{fig:ui}
}

\vgtcinsertpkg

\begin{document}


%\firstsection{Introduction}

\maketitle

\setlength{\abovedisplayskip}{2.5pt}
\setlength{\belowdisplayskip}{2.5pt}

\section{Introduction} 
%for journal use above \firstsection{..} instead
\vspace{-.8mm}
The recovery of underlying causality in observational data is one of the fundamental problems in science. Event sequences are widely collected in the form of \rrv{a} series of \rrv{time-stamped} events across a broad range of applications, such as electronic health records, financial transactions, and web clickstreams. The progression of individual sequences carries rich information \rrv{on} how events are mutually \rrv{effected}. Analyzing collections of temporal event sequences can help analysts extract cause--effect \rrv{relationships between} events, which may be beneficial to various analytical tasks, such as event forecasting and intervention planning. For example, in the medical domain, uncovering the causal relationships residing in sequences of medical records can help doctors understand critical symptoms that indicate a certain disease and promising treatment plans. In digital marketing, exploiting causal factors behind the increase and decrease in sales can \rrv{provide} insights into marketing strategies.

Although \rrv{randomized controlled \rrv{trials}~\cite{cartwright2010randomised} are the gold standard for discovering causality}, conducting such experiments is extremely difficult \rrv{and costly}. Therefore, causal analysis approaches have been developed for inferring causalities from modeling cause--effect relationships in observational data~\cite{pearl2009causal,shimizu2014lingam,malinsky2018causal}. In temporal data, the discovery of causal relationships is commonly based on the theory of Granger causality~\cite{granger1969investigating}, which is defined regarding the predictability and the temporal ordering of events. There is an extensive \rrv{amount} of research that focuses on learning Granger causality in event sequence data, including those based on graphical modeling~\cite{lozano2009grouped,peters2013causal}, Hawkes processes~\cite{xu2016learning, achab2017uncovering}, and deep neural networks~\cite{nauta2019causal}. While these techniques have demonstrated their capabilities in identifying some reliable causal relations, many of them rely on rather general presumptions of the data distributions, which \rrv{may} fail to encode \rrv{a} sufficient amount of domain-specific knowledge~\cite{chen2011data,liu2018steering}. In addition, the high complexity of causal models can lead to a lack of sufficient interpretability and explainability to support decision-making. 

Several recent studies have proposed \rrv{analyzing} causality through visual analytics, attempting to compensate for the deficiency of automatic causality analysis methods by bringing in human supervision~\cite{wang2015visual,wang2017visual}. They utilize a set of visualizations and interactive tools to help human experts investigate and examine causal analysis \rrv{results} obtained from the model. However, these methods are mainly designed for non-temporal multivariate data with \rrv{a} limited number of variables. They are generally not applicable to temporal event sequences, as unique characteristics of event sequence data pose several special \rrv{challenges}. First, event sequence datasets often contain a large variety of event types~\cite{gotz2014decisionflow,guo2018visual}. 
% For example, electronic health records can include up to hundreds and thousands of unique diagnostics, treatment, and lab test codes. 
This high dimensionality of event sequence data can significantly increase the complexity of the causal analysis result. 
% For example, the causal graph may become large in scale and complex in causal structures, preventing efficient exploration of the event causalities. 
Second, sequences of various event types occurring in different orders \rrv{lead to a} high heterogeneity between individuals~\cite{guo2017eventthread}. This hinders \rrv{the extraction and summarization of} common causal patterns in raw event sequences, resulting in \rrv{difficulties} in \rrv{the interpretation and verification of} the causality analysis \rrv{results}.

We introduce~\name to address the aforementioned challenges: (1) the incorporation of human knowledge, (2) \rrv{the} lack of interpretability and explainability of automatic causality analysis, and (3) the temporal complexity specified in causality analysis of event sequence data. \name is an integrated visual analytics prototype designed for analyzing causalities in event sequence data. Concretely, we recover \rrv{the} Granger causality of events within a collection of event sequences based on Hawkes process modeling. To address the first challenge, we further enhance the causality analysis algorithm with a user-feedback mechanism that is able to leverage sparse corrections from human experts to update the entire causal model. Moreover, we \rrv{introduce} a set of visualizations and interactions for exploring, interpreting, and verifying complex causalities in high-dimensional and heterogeneous event sequences to address the second and third challenges. We \rrv{quantitatively evaluate the ability of the user-feedback mechanism to improve the performance of automatic causality analysis, and present case studies to demonstrate} the utility of our visual analytics prototype. The major contributions of this paper are as follow:
% To support human-in-the-loop visual analytics, we further enhance the causality analysis algorithm with a user-feedback mechanism that is able to leverage sparse corrections from human experts to update the entire causal model, so that the causalities derived from the model can reach a higher consistency with human knowledge and achieve better model performance. We present a set of visualizations and interactions that are specially designed for exploring and verifying complex causalities in high-dimensional and heterogeneous event sequences. 
% More concretely, to address the first challenge, we design the system to support bottom-up exploration of the causalities by hierarchical causal structures so that users can focus on relatively small number of events and causal relations at a time. We also incorporate a layout algorithm in the system to emphasize the causal structures (e.g., chains and circles) in the graph. To support users interpreting and examining causalities as required to address in the second and third challenge, we design a flow-based visualization that aggregates sequences by whether they accord with the causal relation under examination. The system also provides a summarization of the causal patterns to help users discover potential causal links. 
% To summarize, the major contributions of this paper are as follows:
\begin{itemize}[leftmargin=\parindent]
\vspace{-2mm}
\setlength\itemsep{-.6mm}
\item \textbf{System.}
We introduce an interactive visual analysis prototype \rrv{that} supports a workflow of exploration, verification, and comparison of causalities in event sequence \rrv{datasets}. To address the exploration difficulty introduced in the third challenge, the system integrates interactions for bottom-up exploration of complex causal \rrv{graphs} to enhance \rrv{the} efficiency in causal exploration. The system also enables users to interpret and examine the validity of causal relations from raw event sequences so as to \rrv{meet} the second challenge.

\item \textbf{Algorithm.}
We design a user-feedback mechanism to enhance \rrv{the} causality analysis algorithm in order to \rrv{address} the first challenge. It is able to transfer user corrections on the automatically generated causal relations to the causal model so that the model can be iteratively refined to better accord with users' domain knowledge.

\item \textbf{Visualization.} 
We design a set of novel visualizations to display event causalities and summarize causal patterns in raw event sequences. This targets at resolving the difficulty in exploring and interpreting causalities in event sequences as introduced in the third challenge. \rrv{In particular}, we employ a causal graph to display causal relations and design a layout algorithm to better reveal causal structures (i.e., causality chains, circles). We also employ a flow-based visualization with an optimized layout for aggregating raw event sequences and \rrv{showing} how sequences progress among key events in the causal graph. 

% \item \textbf{Evaluation.} We report results from two different studies: a quantitative study justifying the capability of the user-feedback mechanism in improving the performance of automatic causality analysis, and a qualitative case study with medical domain experts assessing the utility of our proposed visualizations and visual analytics prototype. The results have shown that the goodness-of-fit and accuracy of the model can be iteratively improved with efficiency when given sparse user feedback, and the system can help medical experts discover and examine useful causal relations.
\end{itemize}

\vspace{-3mm}
\section{Related Work}
\vspace{-0.5mm}
Causal modeling is an active research area with extensive literature. Depending on the types of analyzed data, existing techniques can be broadly categorized into methods for independent and identically distributed (i.i.d.) data and non-i.i.d. data~\cite{guo2018survey}. Temporal event sequences as a special type of non-i.i.d. data require distinct causal modeling algorithms that comply with the \q{temporal precedence} assumption~\cite{holland1986statistics}.
%Mill's definition
In this section, we summarize prior research that is most relevant to our work, including causal modeling techniques specifically designed for event sequence data, and visual analysis techniques developed to facilitate causal analysis. 

\vspace{-1.5mm}
\subsection{Causal Modeling for Event Sequences}
\vspace{-0.5mm}
The progression of successive time-stamped events can carry a great deal of information about the underlying causal mechanism. In this context, many approaches have been developed to recover the mutual causation of events, which mainly includes graphical modeling methods, Hawkes-based techniques, and deep learning approaches. 

Graphical causal models, such as Peter \& Clark (PC) and Functional Causal Model (FCM)~\cite{spirtes2000causation, pearl2009causal}, are well-recognized causal discovery methods originally developed for non-temporal multivariate data. A number of papers attempt to extend typical graphical models to handle temporal data by incorporating an additional restriction on the temporal ordering of cause and effect. For example,  TiMINo~\cite{peters2013causal} and VAR-LiNGAM~\cite{hyvarinen2008causal} enrich the causal equation in FCM with time lags of causal relationships. Similarly, PCMCI~\cite{runge2017detecting} and tsFCI~\cite{entner2010causal} adapt typical conditional independence testing in \rrv{the} time-lagged correlation analysis. Graphical modeling techniques mostly require prior assumptions on the causal relationships, based on which the algorithm searches and verifies true causalities. In many domains, however, the lack of such assumptions and \rrv{a} large number of event types become serious impediments to the application of these methods. 

Another direction of research is based on the theory of Hawkes processes~\cite{hawkes1971point}, which corresponds to an autoregressive event sequence modeling technique that captures the self-excitation and mutual-excitation of events. This underlying principle of Hawkes processes has elicited a group of studies that attempt to recover causal relationships of events over a period of time from their intensity of influence inferred from the model. Eichler et al.~\cite{eichler2017graphical} apply the concept of Granger causality to \rrv{Hawkes} processes using \rrv{a} least squares estimation of the impact function. Xu et al.~\cite{xu2016learning} advance this technique with a set of regularizers to improve the robustness and computational complexity of the model. Unlike \rrv{graphical-modeling-based methods}, which merely estimate the causal relationships between events, Hawkes-based techniques are able to calculate the change of causal strength \rrv{within} each pair of events over time. This information may \rrv{integrate} more causal semantics into the analysis context and result in more interpretable discoveries. 

With deep learning techniques gaining popularity, some recent causal discovery methods attempt to leverage the capabilities of deep neural networks in capturing complex event dependencies. For example, Zhang et al.~\cite{zhang2020cause} utilize \rrv{the neural point processes}~\cite{mei2017neural} based on recurrent neural networks in place of Hawkes processes in causal discovery. Nauta et al.~\cite{nauta2019causal} discover causal relationships and causal delays in temporal data with an attention-based convolutional neural network. Although deep learning approaches generally achieve better accuracy and scalability than graphical modeling algorithms and Hawkes-based causal discovery algorithms, the lack of interpretability poses a great problem for understanding and justifying causal relationships. 

To balance between the informativeness and interpretability of the analysis result, in this paper, we base our work on the \rrv{state-of-art} Granger causality analysis algorithm based on Hawkes processes~\cite{xu2016learning}. In particular, we extend this earlier work to accommodate interactive visual analysis through a user-feedback mechanism that takes users' modifications \rrv{of the initial causal relationships and updates the model} accordingly. Together with the interactive visual interface, our method \rrv{leads to} more accurate and comprehensive causal findings.

\vspace{-1.5mm}
\subsection{Visual Causality Analysis}
A wide variety of methods have been developed to visualize causality in data analysis. Traditional visualizations, such as directed acyclic graph (DAG) layouts and Hasse diagrams, can be employed to illustrate causality to a certain extent. However, they become inefficient as an increasing number of variables may introduce more edge crossings. Elmqvist et al. propose two visual methods, Growing Squares~\cite{elmqvist2003growing} and Growing Polygons~\cite{elmqvist2003causality}, which enhance node representations in DAGs with color-coded squares and polygons to provide an overview of influences on each event. They also leverage \rrv{animation} to present the temporal ordering of causality. Although both methods are effective in uncovering the causal structure of events, they fail to integrate causal semantics into the graph, which is important for a deeper understanding of the causal relationships. To incorporate additional causal semantics, Kabada et al.~\cite{kadaba2007visualizing} introduce a set of animations following Michotte’s rules of causal perceptions~\cite{michotte1963causalite} to intuitively illustrate causal strength, amplification, dampening, and multiplicity. Recent studies put more effort \rrv{into} integrating automatic causal analysis algorithms and causality visualizations into a visual analytics system to facilitate interactive causal analysis and reasoning. Chen et al.'s~\cite{chen2011data} visual causal analysis system aims to provide hypothesis generation and evaluation and support decision-making, which leads to a number of visual analytics systems designed to support interactive analysis of data correlation and causation. For example, Zhang et al.~\cite{zhang2014visual} utilize a force-directed graph layout to present the correlation between numerical and categorical variables in multivariate data. ReactionFlow~\cite{dang2015reactionflow} aims to support \rrv{a} better understanding of causal relationships between proteins and biochemical reactions in biological pathways. It organizes the causal pathways into a flow-based structure to emphasize the downstream and upstream of the causal relationships. To include domain knowledge, Wang and Mueller~\cite{wang2015visual} present an interactive visual interface that allows analysts to edit and verify causal links according to their domain expertise. They further extend this work with a path diagram visualization to better expose causal sequences of the variables~\cite{wang2017visual}.

Despite the extensive visual analytics approaches for analyzing causalities, \rrv{most of the existing techniques focus} on non-temporal multivariate data and methods for analyzing causal relationships in temporal event sequences still remain deficient. \x{Most relevant to our work is the visual analytics framework introduced by Lu et al.~\cite{lu2017visual}, which annotates critical changes \rrv{in} media topic volume with causalities of media events. Our work focuses on extracting accurate causal relationships \rrv{between} events from \rrv{a} general event sequence dataset and assisting analysts in making interpretable causal discoveries.} 
% Our work, however, aims at addressing particular analytics tasks and challenges in analyzing causalities in event sequences. The system is developed to support efficient exploration, verification, and comparison of complex event causation through a set of specifically designed visualizations and interactions.
% When dealing with event sequences, three major challenges need to be specifically tackled. First, temporal nature of event progressions adds additional causal semantics, such as causal delays and causal durations, into the causation of events. This in turn, raises the bar for extracting causal causations in event sequences. Second, the high dimensionality of events and the latent structure of hierarchies in event categories add complexity to the causal graph. This requires a dedicated graph layout mechanism to handle the causal complexity. Lastly, as mentioned earlier, the complexity of temporal event sequence data leads to difficulty in summarizing and investigating event sequence collections. This further raises issues in interpreting and verifying event causation, which we aim to address in this paper.

\vspace{-1mm}
\section{Requirement Analysis and Approach Overview}
\label{sec:overview}
% The goal of \name is to extract accurate causal relationships among events from complex event sequence dataset and assist analysts in making interpretable causal discoveries. 
Prior to the development of \name, we had a thorough discussion with experts in the medical domain on the specific analytical tasks and challenges of analyzing causalities in electronic health records. \rrv{By including design principles from} previous visual causality analysis techniques, we identify a set of design requirements: 
% In each design requirement, we provide detailed descriptions of the problems we aim to address, and discuss general design schemes of the solution. 
\begin{itemize}
\vspace{-2mm}
\setlength\itemsep{-.8mm}
    \item[\textbf{R1.}]\textbf{Extract key events and subgroups for analysis.} Real-world event sequence datasets usually contain \rrv{a} large number of divergent progression patterns and irrelevant event types, which may introduce a lot of noise in causal analysis. The system should, therefore, allow users to query sequence subsets that follow \rrv{a} similar progression context and select key event types to ensure the performance of the causal model.
    % Real-world event sequence datasets, such as electronic health records, can contain up to thousands of sequences and hundreds of event types. This large number of divergent progression patterns and irrelevant event types can introduce a lot of noise in causal analysis. The system should, therefore, allow users to query sequence subsets that follow similar progression context and select key event types to ensure the performance of the causal model. 

    \item[\textbf{R2.}]\textbf{Ensure efficiency in causal exploration.} The high dimensionality of event sequence data can result in a large and complex causal graph that is hard to investigate as a whole. 
    % Specifically, our medical experts suggested that:~\q{The causality in medical dataset can be very complex, with a vast number of complications and medicines being interlinked.} 
    To cope with this issue, the system should incorporate interactive approaches to improve efficiency in exploring causalities.
    
    \item[\textbf{R3.}]\textbf{Enhance \rrv{the} interpretability of causal relationships.} 
    The lack of interpretability is an inherent issue of machine learning models, which also exists in the context of causal analysis~\cite{stokes2017study}. 
    % Our medical experts also commented that:~\q{It is hard to believe and apply the analysis result concluded from automatic analysis models. We need observable evidence from the data to enhance our confidence on the result.} 
    The system should, therefore, provide explanations from underlying data by demonstrating corresponding sequences in the dataset that follow particular causal patterns.
    
    \item[\textbf{R4.}]\textbf{Support identification of spurious causalities.}
    The theory of Granger causality exploits the association of event variables under the restriction of temporal precedence~\cite{granger1969investigating}. However, temporal precedence alone is sometimes not sufficient for establishing true cause--effect relationships~\cite{eichler2013causal}. Hence, the system should support identifying spurious causalities from the causality analysis result.
    
    \item[\textbf{R5.}] \textbf{Incorporate human knowledge in \rrv{the} causal model refinement.} 
    Automatic causal analysis algorithms are generally not capable of \rrv{including an} adequate amount of human knowledge~\cite{chen2011data}. For example, doctors are required to follow medical guidelines \rrv{that} contain verified causalities and restrictions of medical treatments that are not included in the model assumption. 
    % It is also considered beneficial in prior literature~\cite{wang2015visual, wang2017visual} to incorporate human knowledge into causal analysis.
    Consequently, the system should allow users to modify \rrv{the} model output and incorporate user feedback into model refinement.
    
    \item[\textbf{R6.}]\textbf{Provide diagnostic measures on model quality.}
    Bringing human supervision \rrv{into} causal model refinement may introduce user biases in the analysis result~\cite{wall2018four}. To guard against the potential negative effect of the causal model from biases, it is necessary to support objective model diagnostics mechanisms to guide user refinements on the model output.
    
    \item[\textbf{R7.}]\textbf{Allow comparison of causalities for different subgroups.} Causal relations inferred from different groups of sequences can vary dramatically. For example, in \rrv{the} medical scenario, patients may have different applicable medicines due to different symptoms. Comparing causalities of different \rrv{cohorts} can help doctors make personalized treatment plans. To this end, the system should allow comparing causalities in different subgroups of sequences.
    % As suggested by our medical experts, causal relations inferred from different cohort can vary dramatically:~\q{Patients may have different symptoms and applicable medicines according to various physical conditions,} as they suggested:~\q{comparing causalities in different cohort can help us making personalized treatment plans.} To this end, the system should allow comparing causalities in different subgroups of sequences.
\end{itemize}
\begin{figure}[t]
    \centering
    \includegraphics[width=\columnwidth]{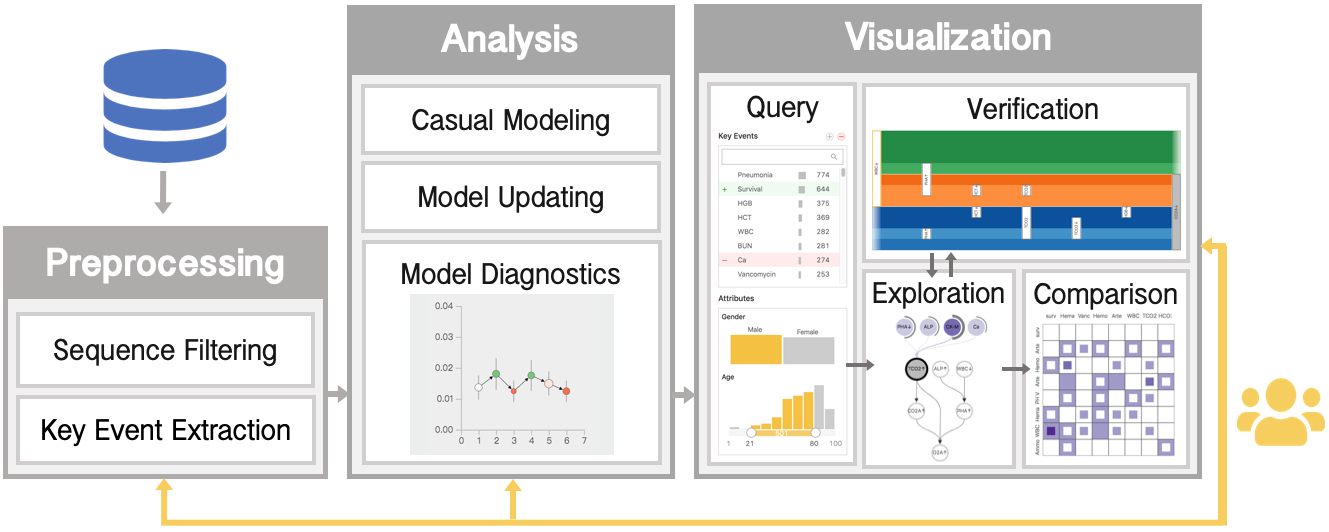}
    \vspace{-6mm}
    \caption{The framework of the \name system, including a data preprocessing module, a causal analysis module, and a visualization module. }
    \label{fig:overview}
    \vspace{-6mm}
\end{figure}

\vspace{-1.5mm}
Guided by the above design requirements, we developed \name, a \x{web-based} visual analytics system for recovering causalities in general event sequence \rrv{datasets}. \x{\name uses the open-sourced JavaScript framework \textit{React}. The front-end functionality is achieved by \textit{D3.js}. The back-end causality analysis algorithm is implemented with Python.}
% The \x{framework} of the system is illustrated in Fig.~\ref{fig:overview}, which consists of three key modules: a data preprocessing module, a causal analysis module, and a visualization module. 
The \x{framework} of the system is illustrated in Fig.~\ref{fig:overview}. 
The data preprocessing module is equipped with an efficient query mechanism that allows users to filter a subset of sequences fitting certain criteria and key events of \rrv{interest} to build a causal model~(\textbf{R1}). 
The causal analysis module is primarily responsible for extracting causal relationships between events from the preprocessed dataset and further delivering to \rrv{the} visualization module for visual causality analysis. In addition, the causal analysis module provides a user-feedback mechanism that integrates modifications from users to update the causal model~(\textbf{R5}) with the underlying model diagnostics guaranteeing model quality upon each iteration~(\textbf{R6}). 
The visualization module supports 
% a visual causality analysis workflow of 
\x{the following functionality}: 1) causal exploration, which supports user-driven \rrv{investigations} and edits of the model output~(\textbf{R2}), 2) causal verification, which summarizes causal patterns in original sequences to help with causal interpretation~(\textbf{R3}) and guide model refinement~(\textbf{R4, R5}), and 3) causal comparison, which allows users to compare causalities of different queries~(\textbf{R7}).

\vspace{-1mm}
\section{Causal Discovery from Event Sequences}
\label{sec:causal_analysis}
\begin{figure*}[t!]
    \centering
    \includegraphics[width=\linewidth]{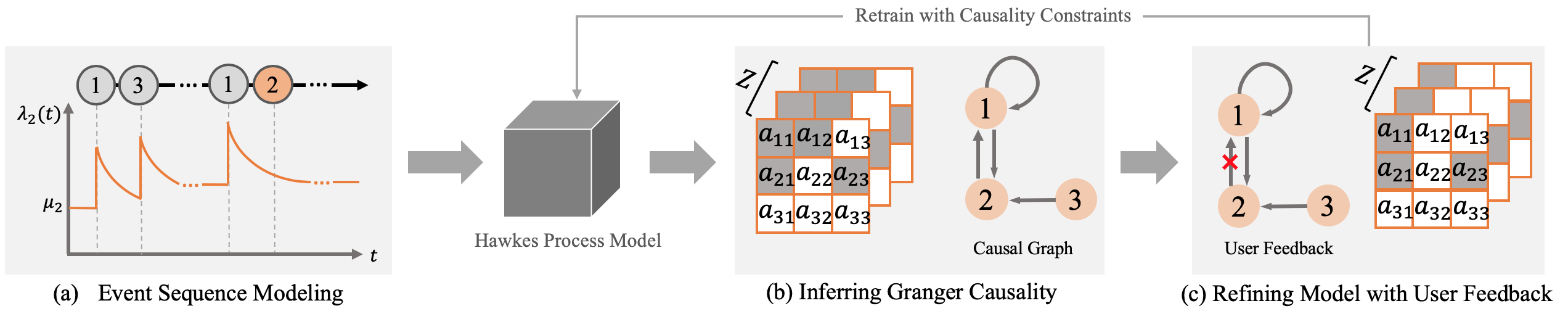}
    \vspace{-7mm}
    \caption{The pipeline of causality analysis algorithm. The algorithm consists of three key steps: (a) training a Hawkes process model to fit the event sequence dataset, (b) inferring impact factors $a_{vv\prime}$ between two events from the trained model to generate initial causal relations, and (c) incorporating users' modifications to retrain the model with a constrained objective.}
    \vspace{-5.5mm}
    \label{fig:algo}
\end{figure*}
This section introduces the causality analysis algorithm for extracting Granger causality from event sequence datasets. Figure~\ref{fig:algo} gives an overview of the causal analysis pipeline, which is composed of three key steps (Fig.~\ref{fig:algo}(a--c)). First, we employ Hawkes processes to model Granger causalities in event sequences using Xu et al's technique~\cite{xu2016learning}. Then, we train the model to fit the data by maximizing the likelihood. The parameters of the trained model are \x{utilized} to infer Granger causalities of event types. Lastly, we enhance the interactivity of the model with a user-feedback mechanism~\x{(\textbf{R5})} \rrv{that} incorporates human knowledge for model refinement. 
% Altogether, this analysis pipeline forms the foundation of the causal analysis module, providing an initial causal analysis result and enabling iterative refinement of the initial result according to human knowledge. 

\vspace{-1.7mm}
\subsection{\x{Background} of Hawkes Processes}
\label{sec:hawkes}
Granger causality is capable of characterizing causality in temporal data according to incremental predictability: if the occurrence of an event $B$ enhances the predictability of \rrv{an} event $A$, event $B$ Granger causes event $A$. Hawkes processes~\cite{hawkes1971point} build a statistical model that describes the triggering patterns between events. The assumption behind Hawkes processes is similar to the theory of Granger causality \rrv{in} the context of event sequence data, which presumes that the occurrence of an event may increase the probability of occurring another event in the near future~\cite{mei2017neural}. This consilience makes Hawkes processes particularly suitable for learning Granger causality in event sequences. Therefore, in the first step, we leverage the Granger causality analysis method of Hawkes processes proposed by Xu et al.~\cite{xu2016learning} to establish our analysis model.
% In Hawkes processes, the conditional intensity (i.e, number of event occurrences per time unit) of an event is determined by two factors: a constant baseline intensity independent of other events, and a time-varying intensity stimulated by the occurrence of other events and decays over time. The conditional intensity of an event, by its definition~\cite{liniger2009multivariate}, can imply the occurrence probability of the event over time. 
% For example, as shown in Fig.~\ref{fig:algo}(a), $\lambda_{2}(t)$ represents the conditional intensity of event $2$ over time with a constant baseline value $\mu_{2}$, which is stimulated on the occurrence of events $1$ and $3$ where the effect of stimulation gradually decays to the baseline. 
\x{Given a collection of event sequences with $V$ types of events, the occurrence probability of event $v\in \{1,\dots,V\}$ at time $t$ can be inferred from its conditional intensity (i.e., number of event occurrences per time unit), $\lambda_{v}(t)$, which is formally defined as:}
% Given a collection of event sequences with $V$ types of events, the conditional intensity $\lambda_{v}(t)$ of event $v\in \{1,\dots,V\}$ at time $t$ can be formally defined in the following:
\begin{align}
    \lambda_{v}(t)=\mu_{v}+\sum_{v^{\prime}=1}^{V} \int_{0}^{t} \phi_{v v^{\prime}}(r)\,\mathrm{d}N_{v^{\prime}}(t-r)
\end{align}
where the first term $\mu_{v}$ is \x{a constant} baseline intensity, and the second term indicates the increase of intensity brought by the excitation of all historical events on event $v$. Specifically, $N_{v^{\prime}}(t-r)$ is the number of events $v'$ before time $t-r$, and $\phi_{v v^{\prime}}(r)$ is a time-varying impact function that captures the influence of historical event $v'$ on event $v$:
\begin{align}
    \phi_{v v^{\prime}}(t)=\sum_{z=1}^{Z} a_{v v^{\prime}}^{z} \kappa_{z}(t)
\end{align}
In particular, the impact function incorporates a linear combination of a set of $Z$ Gaussian sampling functions $\{\kappa_{z}(t)\}_{z=1,\dots,Z}$ to simulate the decaying influence of event $v'$ on event $v$. $Z$ is set as the minimum value according to Silverman's rule of thumb~\cite{Silverman86}, which limits the maximum brandwidth for Gaussian sampling that is calculated based on the time duration between two events and the number of times that two events co-occur in one sequence.  $\bm{a}_{vv^{\prime}}=[a_{vv^{\prime}}^{1},\dots,a_{vv^{\prime}}^{Z}]^\top$ are the impact coefficients, which indicate the level of stimulation effect caused by event $v^\prime$ on event $v$. 

% How is $N$ determined?
% In this way, $\lambda_{v}(t)$ is estimated via a Gaussian-based kernel density estimator with bandwidth controlled by $h_{vv^{\prime}} = \left((4 \hat{\sigma}_{vv^{\prime}}^{5})/(3 \sum_{i} M_{i_{vv^{\prime}}})\right)^{0.2}$ according to Silverman's rule of thumb \cite{Silverman86} where $\hat{\sigma}_{vv^{\prime}}$ is the stand deviation of time duration between event-$v$ and event-$v^{\prime}$ and $\sum_{i} M_{i_{vv^{\prime}}}$ is the number of the event pairs $(v^{\prime},v)$ in dataset. Thus, we can set the optimal $N = Min(T_{vv^{\prime}}/h_{vv^{\prime}})$ where $T_{vv^{\prime}}$ is mean value of the time duration between event-$v$ and event-$v^{\prime}$.

\vspace{-1.7mm}
\subsection{Learning Granger Causality}
\label{sec:lgc}
%first introduce (4) and break down to (3)
In the second step, we search over the parameter space to fit the Hawkes processes to the sequence collection. The parameters include the base intensity of all events $\bm{\mu} = [\mu_{v}]_{v=1,\dots, V} \in \mathbb{R}^{V}$ and the impact coefficients $\bm{a} = [a_{vv^{\prime}}^{z}]_{v,v^{\prime}=1,\dots,V}^{z=1,\dots,Z} \in \mathbb{R}^{V\times V\times Z}$ for each pair of events $(v,v')$. We formulate the training objective as follows:
\begin{align}
\label{objective}
    {\rm argmin}_{\bm{\mu},\bm{a}} \quad - L + \alpha \sum_{v, v^{\prime}}\left\|\bm{a}_{v v^{\prime}}\right\|_{2}
\end{align}
where the first term is the negative log-likelihood of the \x{Hawkes process} on the sequence dataset \cite{hawkes1971point}. \x{Given a collection of event sequences $S = \{s_{i}\}_{i=1,\dots,I}$, where each sequence $s_{i} = \{(v_{m}^{i},t_{m}^{i})\}_{m=1,\dots,M_{i}}$ is a series of $M_i$ event--time pairs with $v_{m}^{i} \in \{1,\dots,V\}$ and $t_{m}^{i}$ representing the event type and timestamp of the $m$-th event respectively, the log-likelihood $L$ can be expressed as follows:}
\begin{align}
    L=\sum_{i=1}^{I}\left\{\sum_{m=1}^{M_{i}} \log \lambda_{v_{m}^{i}}\left(t_{m}^{i}\right)-\sum_{v=1}^{V} \int_{0}^{T_{i}} \lambda_{v}(r) d r\right\}
\end{align}\\
% $L$ can be interpreted as the probability of occurrence for each event in each sequence. The first term is a summation of estimated intensities for all events in sequence $s_i$, and the second term is the intensities of all event types in the continuous period of time from the beginning to the end of the sequence $T_{i}$. When maximizing the likelihood, the first term guarantees that the estimated intensities of all events in sequence $s_i$ fit well with the observations, and the second term is subtracted to ensure that the temporal distribution of intensities is consistent with the time of event observations in the sequence.
The second term of the training objective is a group-lasso regularizer which ensures \rrv{that} the inferred impact coefficients are interpretable \x{by the theorem of Eichler et al.~\cite{Eichler2016GraphicalMF}}. 
\x{According to the theorem, the Granger causality between event types can be directly inferred from the impact coefficients $\bm{a}_{vv^{\prime}}=[a_{vv^{\prime}}^{1},\dots,a_{vv^{\prime}}^{Z}]^\top$, and events $v$ and $v^\prime$ have no causal relationship only if $a_{vv^{\prime}}^{z}=0$ for all $z \in \{1,\dots,Z\}$. }The hyperparameter $\alpha$ controls the influence of the regularization term.
% The addition of the group-lasso regularizer ensures that $Z$ impact \x{coefficients} for each pair of events are either excluded (i.e., set to 0) or included together when updating the parameters. The hyperparameter $\alpha$ controls the influence of the regularization term.

The objective function is optimized by applying an EM-based algorithm \cite{Lewis2011RESEARCHAA}, and the learning result $\bm{a} = [a_{vv^{\prime}}^{z}]_{v,v^{\prime}=1,\dots,V}^{z=1,\dots,Z}$ captures the causal relationships and causal strengths between events. \x{We define the causal strength of the causality $v^{\prime} \rightarrow v$ as follow}:
\begin{align}
    {\rm Strength}_{v v^{\prime}}=\frac{1}{Z} \sum_{z=1}^{Z} a_{v v^{\prime}}^{z}
\end{align}
Thus, we can obtain a directed causal graph $G(\mathcal{V},\mathcal{E})$ whose edges $\mathcal{E}=\{v^{\prime} \rightarrow v\}$ are weighted by the causal strength ${\rm Strength}_{vv^{\prime}}$.

\vspace{-1.7mm}
\subsection{Updating Causality with User Feedback}
\label{sec:user-feedback}
%强调之前的算法不支持feedback
%We extend..(x) -> we further designed..
To incorporate human knowledge in causality analysis~\textbf{(R5)}, we further designed a user-feedback mechanism that is able to make refinements on the model according to user inputs. In particular, the user can modify the causal graph from the visual interface by preserving authentic causal relations and deleting spurious ones. Based on the user's modifications, a new causal graph $\hat{G}(\hat{\mathcal{V}},\hat{\mathcal{E}})$ is generated, and the model can be updated automatically by optimizing a new objective function:
\begin{align}
    & {\rm argmin}_{\bm{\mu},\bm{a}}  \quad - L + \alpha_{u}\sum_{v, v^{\prime}}\left\|\bm{a}_{{v v^{\prime}}^{(\hat{G})}}\right\|_{2} \\
    & {\rm s.t.} \quad \bm{a}_{v v^{\prime}} = \bm{0} \quad {\rm for} \quad (v^{\prime} \rightarrow v) \notin \hat{G} \nonumber
\end{align}
where $L$ is the log-likelihood of Hawkes process, $\alpha_{u}$ is the control hyperparameter, and $\sum_{v, v^{\prime}}\left\|\bm{a}_{{v v^{\prime}}^{(\hat{G})}}\right\|_{2}$ is the user-specified regularizer:
\begin{equation}
    \bm{a}_{{v v^{\prime}}^{(\hat{G})}}=\left\{\begin{array}{cl}
\bm{0} ; &  {\rm if} \quad (v^{\prime} \rightarrow v) \quad {\rm is} \quad {\rm confirmed} \\
\bm{a}_{v v^{\prime}} ; & {\rm otherwise}
\end{array}\right.
\end{equation}
Specifically, if a causal relation is removed by the user, the constraint of Equation (6) ensures that the model parameters are optimized toward setting the corresponding impact factor as 0, so that the updated causal model aligns with user feedback. If a causal relation is confirmed by the user, the updates of the corresponding impact factor can be liberated from the group-lasso regularizer according to Equation (7). This aims to prevent the impact factor from being set as 0 by the regularizer. After refining the model, the causal graph will be redrawn based on the updated parameters $\bm{a} = [a_{vv^{\prime}}^{z}]_{v,v^{\prime}=1,\dots,V}^{z=1,\dots,Z} \in \mathbb{R}^{V\times V\times Z}$. The user can investigate the updated causalities and iteratively make modifications until the analysis result is satisfactory.

\textbf{Computational complexity.} The computational complexity of the causality analysis algorithm is $O(ZV^2n^3)$ per training iteration, which is the same for both the initial computation and the update of the causality on user-feedback. It depends on three data attributes: the number of the sampling functions $Z$, the number of event types $V$, and the number of occurrences \rrv{of} all events in the dataset $n$. \x{We implemented the causality analysis model with Python using \rrv{the} NumPy package, which is able to compute the causality between events in parallel. \rrv{Running times} under different data sizes \rrv{are} reported in Section~\ref{sec:dis}.}
% $n={\rm max}_{_{1 \leq i\leq V}}n^{i}$, where $n^{i}$ is the number of occurrences of event $i$ in the dataset.

\section{Visual Causality Analysis}
% We design \name to support visual causality analysis through a multi-view user interface driven by the interactive framework introduced in Section~\ref{sec:overview}. 
In this section, we first introduce the main components in the \name interface. Then, we describe the details for each system functionality. 
% following the visual causality analysis workflow, which includes three processes: causal exploration, causal verification and causal comparison. 
% \x{Finally, we present a set of interactions that are designed to boost the process visual causality analysis.}
% The analysis is initiated in the causal exploration process as the user investigate the causal relations derived from the causality analysis model. Then, the user can verify the causalities according to raw event sequences and make corrections to refine the causality analysis model in the causal verification process. After confirming the causalities for a number of data subsets, the user can move to causal comparison process and compare the causal relations of either two subsets. 

% \begin{figure*}[t!]
%     \centering
%     \includegraphics[width=\linewidth]{figures/user_interface.png}
%     \caption{The user interface of \name system.}
%     \label{fig:ui}
% \end{figure*}

\subsection{User Interface}
The user interface of \name is composed of six key views. 
\x{The left panel includes \rrv{the} \textit{query view}~(Fig.~\ref{fig:ui}(1)), allowing the user to select a dataset and filter sequences from the database for analysis~(\textbf{R1}), and a \textit{sequence list view}~(Fig.~\ref{fig:ui}(2)), showing the profile of each individual sequence determined by the query result. }
% The analysis starts with the \textit{query view}~(Fig.~\ref{fig:ui}(1)), which enables the user to select a dataset and filter sequences from the database for analysis~(\textbf{R1}). The \textit{sequence list view}~(Fig.~\ref{fig:ui}(2)) displays the profile of each individual sequence determined by the query result. 

\x{Views in the middle are designed to support causal exploration and verification. The \textit{causal model view}~(Fig.~\ref{fig:ui}(3)) suggests potential causal relationships using a node-link causal graph, allowing users to investigate the causalities and make updates on the causal model after verifying the causal relations~(\textbf{R2, R4, R5}). The \textit{causal sequence view}~(Fig.~\ref{fig:ui}(4)) facilitates causal verification by showing the causal patterns in raw event sequences using a flow-based visualization. We separate these two functionality of our system into two visualizations so that the user can turn to raw event sequences only when needed. The user can verify causal relations according to their domain expertise without investigating the \textit{causal sequence view}, or leverage aids from raw event sequences to assist causal reasoning. In addition, two different hierarchical layouts for the \textit{causal model view} and the \textit{causal sequence view}, emphasizing causal structures (i.e., causality chains and circles) and temporal ordering of events, respectively. To guide iterative updates of the causal graph, the \textit{model diagnostics panel}~(Fig.~\ref{fig:ui}(d)) shows incremental changes of the model quality~(\textbf{R6}). }

% The \textit{causal model view}~(Fig.~\ref{fig:ui}(3)) presents the causal relationships of events with a causal graph for users to explore and modify~(\textbf{R2, R4, R5}). We also display the incremental changes of the model quality for each modification with a \textit{model diagnostics panel} in the \textit{causal model view}~(\textbf{R6}).
% The \textit{causal sequence view}~(Fig.~\ref{fig:ui}(4)) summarizes the causal patterns from raw event sequences with flow-based visualization to facilitate causal verification and interpretation~(\textbf{R3, R4}).

\x{Views on the right include \rrv{the} \textit{analysis history view}~(Fig.~\ref{fig:ui}(5)), \rrv{which} stores user queries and the causal analysis result of the corresponding sequence subdivision from which users can select two subgroups for comparison and \rrv{the} \textit{causal comparison view}~(Fig.~\ref{fig:ui}(6)), showing the differences between causal graphs inferred from two subgroups of sequences through a matrix-based visualization~(\textbf{R7}). } 

\vspace{-1.6mm}
\subsection{Causal Exploration}
\label{sec:causal_dis}
Real-world event sequence datasets are generally large and heterogeneous, containing \rrv{many} event types. This characteristic can lead to great challenges in visualizing and exploring complex event causalities. \x{Therefore, we designed our system to enable flexible data selection, display causalities with intuitive visualizations, and provide efficient interactions to allow exploring causalities incrementally.} 

% Specifically, ~(\textbf{R1}). The \textit{causal model view} displays event causalities through bottom-up interactive expansion of the causal graph, so as to relieve users from the burden of investigating a large and complex causal graph.  To cooperate the causal discovery process in the front-end, \name is equipped with a \textit{query view} for flexible data selection and a \textit{causal model} view for users to investigate and modify causal analysis result. 

\textbf{Select sequences for analysis~(\textbf{R1}).} 
\rrv{The} \textit{query view} for filtering homogeneous subsets from a large collection of event sequences \rrv{ensures} a high quality of causality analysis. The user can choose a dataset from the drop-down list and filter sequences based on the occurrence of key events and the attribute of records. The \textit{key events panel} displays the list of all event types in the dataset, which allows users to determine event-based query criteria. \x{For example, a doctor may need to filter patients diagnosed with certain diseases or taking specific medicines for analysis.} The event types are ranked by the coverage rate (i.e., the proportion of sequences containing each event), which is visually encoded with the length of a horizontal bar, and the exact number of sequences being covered is labeled at the right. An event search is also provided to help the user navigate the event list. By switching between \img{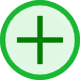} and \img{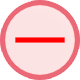} in the \textit{query view}, the user can select events by inclusion criteria or exclusion criteria. The selected events \rrv{are} highlighted with green and red background correspondingly. When executing the query, the system only retrieves sequences that contain all events in the inclusion criteria and no events in the exclusion criteria. The \textit{attributes panel} shows the distribution of records on metadata (e.g., gender and age), allowing the user to filter sequences based on non-temporal attributes. \x{To reduce the negative influence of heterogeneous sequence progression on the performance of the causality analysis, the system further clusters sequences by their progression similarities using the measure proposed by Wongsuphaswat et al.~\cite{wongsuphasawat2012querying}, and retrieve a major cluster of sequences for analysis.} 

\textbf{Visualize and explore causal relationships (R2).} 
After querying the dataset, the causal analysis module generates the causal relationships of all event types in the dataset. The causal relationship is demonstrated in a node-link causal graph, with nodes representing event types and links representing causal relationships pointing from the cause to the effect. \x{To support investigating causal relations in the graph, we design a layout algorithm to reduce the visual complexity of the graph when facing \rrv{a} large number of event types and complex causal \rrv{structures}.} As suggested in a previous study~\cite{bae2017understanding}, laying out the causal graph in sequential order can facilitate the searching process of root cause and derived effects. Following this finding, we choose to visualize the causal graph using a top-bottom sequential layout. However, causal structures in event sequence data can be more complex than sequential chains of cause--effect relations. It may also include causal \x{circles} (e.g.,~(Fig.~\ref{fig:ui}(f))) or self-exciting causalities that cannot be satisfied by simply applying a sequential layout. Therefore, we devise a layout algorithm that calculates the position $\mathbf{p}_i = (x_i, y_i)$ for each node $i \in \{1,\dots, V\}$ to better illustrate the local structures (i.e., causal chains and circles) in the causal graph. Algorithm~\ref{algo:layout} gives an overview of the key steps in the layout algorithm, which is detailed as follows.

\vspace{-3mm}
\begin{algorithm}
\caption{Causal Graph Layout Algorithm}
\label{algo:layout}
\textbf{Input:} The directed causal graph $G(\mathcal{V},\mathcal{E})$\\
\textbf{Output:} A position $\mathbf{p}_i = (x_i, y_i)$ for each vertex $i$ of $\mathcal{V}$
\begin{algorithmic}[1]
\State Detect circles $C = \{C_n(\mathcal{V}_{c_n}, \mathcal{E}_{c_n})\}_{n=1, \dots, N_c}$ by depth-first search
\State Traverse $G$ by breadth-first search and calculate node depth $d_i$ for each vertex $i$
\For{$i \in \mathcal{V}$}
\State Calculate $y_{i} = (CanvasHeight / MaxDepth )\times d_i$
\EndFor
\State Update $\mathbf{p}_i$ for all vertices by minimizing Equation~(\ref{eqa:layout_obj})
\State Update $x_i$ for all vertices to remove \rrv{node} overlap
\end{algorithmic}
\end{algorithm}
\vspace{-3mm}

Given the causal graph $G(\mathcal{V},\mathcal{E})$, we first use depth-first search to detect causality circles $C = \{C_n(\mathcal{V}^{c}_{n}, \mathcal{E}^{c}_{n})\}_{n=1}^{N_c}$ in the graph. \x{To prevent endless loops when iterating over the graph, we take nodes in the same causality circle as one unit. }
% Node depths are calculated according to their sequential ordering in causal chains, and nodes in a circle unit are assigned with the same node depths as the entrance node of the circle. 
% Nodes in the same causality circle are taken as one unit, so that the causal graph can be laid out following regular sequential ordering of nodes in causal chains.
% Then, we layout nodes on y-axis according to the sequential ordering of nodes in causal chains. 
Then, we transform the causal graph into a minimum spanning tree by traversing the graph using breadth-first search. Each node or causality circle is assigned with a depth $d_i$ indicating their level of hierarchy. \x{Nodes within a circle unit are assigned with node depths same as the entrance node of the circle.} Based on the depths, we determine the $y$-position for each node and causality circle by fitting the hierarchy into \rrv{the} canvas in a top-to-bottom manner. 

To minimize edge crossings between adjacent levels of the hierarchy, we further arrange the position of nodes by minimizing the following objective function:
\begin{equation}
\label{eqa:layout_obj}
    J(x,y)=
    S(x)
    +
    \sum_{n=1}^{N_{c}} \sum_{(v\shortto v') \in \mathcal{E}^{c}_{n}} w_{vv'}^c 
    \left\|\mathbf{M}_{n}(\mathbf{p}_v-\mathbf{p}_{v'})-(\mathbf{q}_{nv}-\mathbf{q}_{nv'})\right\|^{2}
\end{equation}
The first term is the stress function of Stress Majorization~\cite{gansner2004graph} we employ to minimize edge crossings as follows:
\begin{equation}
    S(x)=\sum_{i<j} w_{i j}\left(\left\|x_{i}-x_{j}\right\|-d_{i j}\right)^{2}
\end{equation}
where $x_i$ is the $x$-position of $i$-th node, $d_{ij}$ represents the graph-theoretical distance between nodes $i$ and $j$, and $w_{ij}=d_{ij}^{-2}$ is the normalization constant that prioritizes nodes with small distance. 
The second term in the training objective is a circular constraint for \x{regularizing the shape of causality circles.}
% laying out causality circles in the graph. 
$N_c$ is the number of circular sub-graphs in the causal graph. The positions of nodes in the $n$-th circle, $\{\mathbf{p}_v\}_{v\in \mathcal{V}^{c}_{n}}$, are calculated through an affine transformation $\mathbf{M}_n$ \rrv{that} matches the nodes in the $n$-th circle to a reference shape with equal number of vertices positioned at $\{\mathbf{q}_{nv}\}_{v\in \mathcal{V}^{c}_{n}}$. The optimization goal of the second term is to minimize the difference of distances for any two vertices in the causality circle (i.e., $\mathbf{M}_n(\mathbf{p}_v-\mathbf{p}_{v'})$) and the corresponding vertices in the reference shape (i.e., $\mathbf{q}_{nv}-\mathbf{q}_{nv'}$). Similar to the Stress Majorization, the second term also employs a normalization constant $w^c_{vv'} = \left\|\mathbf{M}_{n}(\mathbf{p}_v-\mathbf{p}_{v'})\right\|^{-2}$ to prioritize closer nodes in the causality circle.
The transformation matrix for each causality circle $\mathbf{M}_n$ is calculated by adapting the circular constraint in Wang et al.~\cite{wang2017revisiting} defined as follows:
\begin{equation}
    {\rm argmin}_{\mathbf{M}_n} \sum_{i \in \mathcal{V}^{c}_{n}} \omega_{i} \left\|\mathbf{M}_n  \mathbf{p}_{i}-\mathbf{q}_{ni}\right\|^{2}
\end{equation}
where $\omega_{i}$ is set to the degree of vertex $i$ for illustrating the circular structure more clearly. In the final step, we tweak node positions on \rrv{the} $x$-axis to remove overlaps~\cite{dwyer2005fast}. \x{The layout algorithm runs at the back-end of our system. The objective function is implemented in Python and optimized using NetworkX and SciPy.}

% \textbf{Show causality on user demand (R2).} 
\x{The complete causal graph of an event sequence dataset can be large and complex given \rrv{a} large number of event types. To relieve the user from investigating and verifying many cause--effect relations at a time (\textbf{R2}), \name incorporates a user-driven causal exploration procedure that allows the user to focus on investigating causal relations related to an outcome event incrementally from \rrv{the} bottom to the top. The intention is to uncover only causal pathways that lead to an outcome event of interest to reduce the number of events and causal relations evolved, and eliminate invalid chains of causal relations before branching out. This procedure starts by adding an outcome event as an initial effect using \img{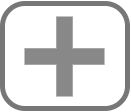} at the top of \rrv{the} \textit{causal model view}. By double-clicking on the effect, the graph expands one layer at the top to show \rrv{the} direct causes suggested by the causal model. As shown in Fig.~\ref{fig:ui}(3), the effect under inspection is colored in \rrv{gray} and the causes are colored by their causal strength. In addition, an outer ring is displayed on each node, and the length represents event coverage (i.e., the proportion of raw sequences that have the cause and effect events appear successively). The user can filter causal relations by causal strength (Fig.~\ref{fig:ui}(b)) and event coverage (Fig.~\ref{fig:ui}(c)). To explore causal chains of the outcome event, the user can continue to expand the graph by iteratively uncovering causes of the topmost events, and stop when the causal chain for the outcome event at the bottom is completed (i.e., no new event is added to the graph). In each iteration, we primarily highlight the newly involved events and causal relations to \rrv{guide} the user's attention \rrv{toward} inspecting them. The user can switch to an overview of the entire graph at any time by clicking on the background of the \textit{causal model view}. }

\begin{figure}[t!]
    \centering
    \includegraphics[width=\columnwidth]{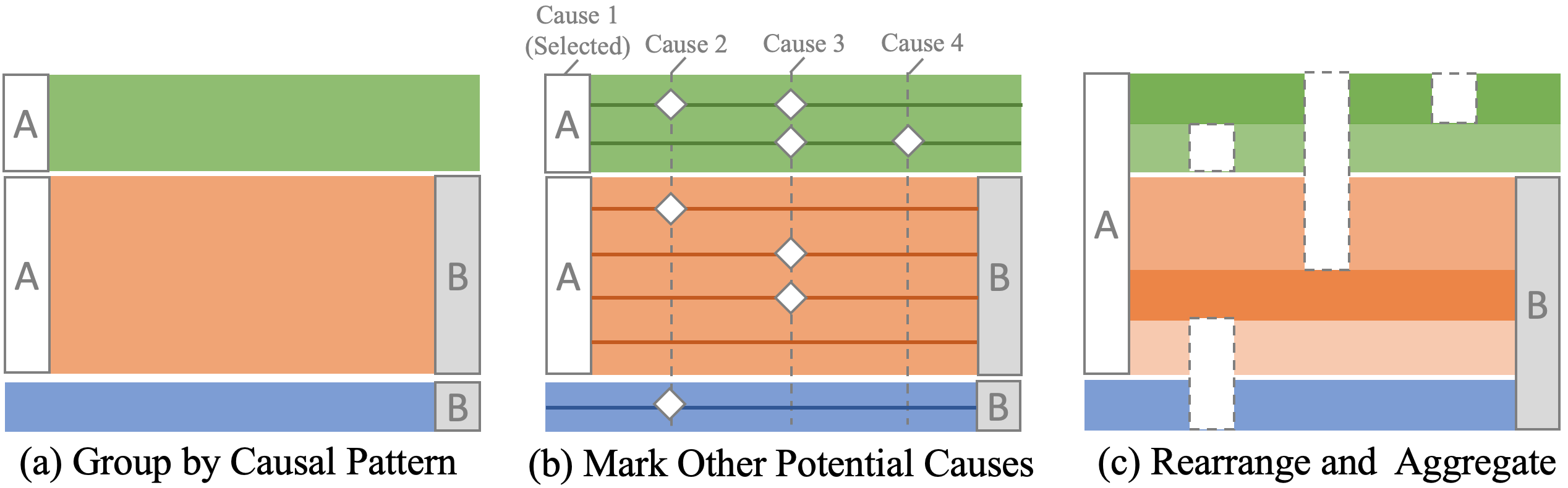}
    \vspace{-6mm}
    \caption{The \textit{causal sequence view} displays causal patterns in raw event sequences related to a selected cause--effect relation (e.g., A\textrightarrow B). (a)~Event sequences are divided into three groups based on the presence of the causes and effects. (b) \x{All other potential} causes are vertically aligned and arranged from left to right in temporal order, and then (c) aggregated to help users examine the validity of the selected cause--effect relation and suggest \x{other possible \rrv{causalities}}.}
    \label{fig:causal_seq}
    \vspace{-4mm}
\end{figure}

\vspace{-1mm}
\subsection{Causal Verification}
\label{sec:causal_verif}
\x{To build a causal model that conforms to objective rules and the user's domain knowledge, \name allows the user to verify the causal relations suggested by the algorithm and update the causal model. In particular, \name displays causal patterns in raw event sequences to support the user \rrv{in} interpreting the causality~(\textbf{R3}) and identifying invalid causal relations~(\textbf{R4}).}  
% To build a causal model that conforms to objective rules and the user's domain knowledge, \name allows the user to verify the causal relationships through visualizing the causal patterns in raw event sequences. 
% Through investigating the causal patterns from original data, the user shall be able to interpret the causality~(\textbf{R3}) and identify invalid causal analysis result~(\textbf{R4}). 
After verifying the causality, the user can modify the causal graph, \rrv{for example,} delete mistaken or add omitted causal relationships, to update the causal model~(\textbf{R5}). Real-time model diagnostics \rrv{are} provided concurrently with user modification to ensure \rrv{high} model quality~(\textbf{R6}). In the following, we introduce the system functionality designed to support causal verification and modification.

\textbf{Visualize causal patterns in raw sequences (R3, R4).} 
To help users interpret and examine the validity of causality analysis results, we associate the calculated causal relations with original data by uncovering the causal patterns in raw event sequences. \x{The user can select a causal relation (e.g., event \textbf{A} causes event \textbf{B}) from the \textit{causal model view} and observe how the sequences progress through the causes and effects from the \textit{causal sequence view}. In particular, there are three categories of relative causal patterns: sequences that go through the cause but never come across the effect afterward (\textbf{A\textrightarrow ?}), sequences that contain both the cause and effect in successive order (\textbf{A\textrightarrow B}), and sequences that have the effect but not the cause before (\textbf{?\textrightarrow B}). To distinguish these situations, we categorize raw sequences into three groups using a flow-based visualization (Fig.~\ref{fig:causal_seq}(a)).} The leftmost and rightmost nodes indicate the cause and effect event colored in white and \rrv{gray}, respectively. The edges between nodes indicate groups of subsequences, colored by the sequence categories. The height of nodes and edges 
is proportional to the number of sequences in the group. \x{We allow users to control causal delays in the subsequences by setting the length of subsequences with a \textit{time delay slider} (Fig.~\ref{fig:ui}(a)).}

\x{The system also displays other potential causes suggested by the causal graph on the subsequences to help the user explore other possible causal relations~(\textbf{R4}) and justify the validity of the \rrv{selected} causal relation. For example, in Fig.~\ref{fig:ui}(4), the raw event sequences are categorized into three groups according to a selected causal relation: \textit{POTA} \textrightarrow \textit{WBC}. To facilitate comparing the selected cause with other potential causes, the system also marked the occurrence \rrv{of} other direct causes besides \textit{POTA} on the subsequences. Specifically, we first mark each potential \rrv{cause} in each individual \rrv{subsequence}. As shown in Fig.~\ref{fig:causal_seq}(b), each line represents an individual subsequence, and the potential causes are anchored in the subsequences. The causes are vertically aligned and horizontally ordered from left to right by their average time of occurrence. Note that we only display one time of occurrence for each cause, as the frequency of occurrence does not affect the validity of causality. In order to simplify the visualization and make it easier to observe the commonness in the occurrence of potential causes, we further reorder subsequences within each category and aggregate common potential causes in adjacent subsequences (as shown in Fig.~\ref{fig:causal_seq}(c)). 
Causes in adjacent subsequences of different \rrv{categories} are also aggregated to further reduce the number of intermediate nodes. In particular, we leverage the reordering algorithm as follows.}

We first calculate the pair-wise similarity of two subsequences $S_i$ and $S_j$ as follows: 
\[
    d\left(S_{i}, S_{j}\right)=\left\|\mathbf{w}\odot(\mathbf{v}_{i}-\mathbf{v}_{j})\right\|  \quad  \mathbf{w}, \mathbf{v}_{i}, \mathbf{v}_{j} \in \mathbb{R}^n
\]
where $n$ is the number of \x{potential} causes. $\mathbf{v}_i$ and $\mathbf{v}_j$ are the one-hot vectors representing the occurrence of \x{potential} causes in subsequences $S_i$ and $S_j$. $v_{i,k} = 1$ if the $k$-th \x{potential} cause \rrv{appears} in $S_i$, otherwise $v_{i,k} = 0$. $\textbf{w}$ is a constant vector that represents the coverage rate of each \x{potential} cause, which prioritizes the aggregation of events that occur in most subsequences. Then, we abstract the sequence ordering problem into a Traveling Salesman Problem (TSP), in which the concept of cities and distances represent subsequences and the pair-wise similarity, respectively. We \rrv{utilize} simulated annealing~\cite{geng2011solving} to search for an accessing order with approximately minimal cost. In this way, subsequences in close proximity can have more \x{potential causes} in common that can be aggregated. 

\x{Our system supports the user to switch among potential causes by either clicking on the nodes in the \textit{causal model view} or the \textit{causal sequence view}. On switching to another potential cause, the corresponding node in the \textit{causal sequence view} will move to the left end with \rrv{a} smooth transition, and the rest of the view will be updated accordingly. The user can also select a causal path (i.e., chains of cause--effect events) by clicking \img{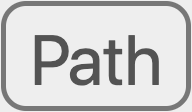} in the \textit{causal model view}, and successively select a series of cause--effect events to emphasize the causal path on the graph. The \textit{causal sequence view} will be updated with the events in the causal path arranged from left to right, and edges showing the progression pattern of sequences flow through the causal path.}

\textbf{Verify and modify causalities (R4, R5).}
The user can determine whether a causal relationship holds true according to the observations in the \textit{causal sequence view} \x{or based on their domain expertise}. For example, if the sequences contain large numbers of \q{A\textrightarrow ?} and \q{?\textrightarrow B} patterns \x{or mostly go through another potential cause, the direct causal relation is not likely to be true. Moreover, we measure the probability that the selected causal relation is valid on a particular \rrv{subgroup} of sequences by calculating the regression likelihood. This probability is encoded by the color saturation of edges, and edges \rrv{with deeper colors} indicate the selected causal \rrv{relationships} generally fit better to the group of sequences.} 
% Moreover, we encode the regression likelihood of the selected causal relationship on different subgroups of sequences into the color saturation of edges. Darker edges indicate the selected causal relationship generally fit better on the group of sequences. 
After investigating \rrv{this statistical} information and incorporating the domain knowledge, the user can determine whether the causal relationship holds true and eliminate spurious causalities. By clicking \img{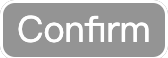} in the tooltip (Fig.~\ref{fig:ui}(e)), a causal relation is confirmed and the corresponding cause event will be colored in gray. After the users finish confirming the causal relationships, they may update the causal analysis model by clicking \img{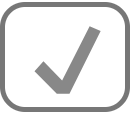}. In response, the causality analysis model will be retrained with \rrv{the} user's feedback of the confirmed causal relations and update the causal graph with the regenerated causality analysis result. The layout of the causal graph is recomputed following Algorithm~\ref{algo:layout}. To make it easier to track nodes in \rrv{the} causal graph before and after the update, we add a stabilization constraint $\sum_{i}\left\|x_{i}-x_{i}^{\prime}\right\|^{2}$ to the original training objective (Equation~(\ref{eqa:layout_obj})) when performing graph updates. The stabilization constraint iterates over common nodes of \rrv{the} causal graph before and after the update and minimizes their change in $x$-position. After updating the causal graph, the user can either continue to explore \x{the causes for topmost confirmed nodes by double-clicking them}
% next-level causalities by double clicking the confirmed nodes, 
or save the final causal analysis result for the queried sequences to the \textit{analysis history view} by clicking \img{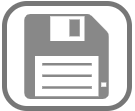} in the \textit{causal model view}.

\textbf{Diagnose the causal model (R6)}.
\x{Every time the user updates the causal analysis model, the \textit{model diagnostics panel}~(Fig.~\ref{fig:ui}(d)) records the change \rrv{of the} overall model quality. This aims to help the user determine the number of iterations to update the graph, which may vary between datasets according to their causal complexity (e.g., lengths of the causal chains and the number of event types). In general, the user can choose to stop adding more iterations when the model shows no significant improvement. The performance of the model is evaluated by three metrics: the regression likelihood of all causal relationships on the queried data, Bayesian Information Criterion (BIC)~\cite{burnham2004multimodel}, and p-value.}
% Every time the user updates the causal analysis model, the \textit{model diagnostics panel}~(Fig.~\ref{fig:ui}(d)) records the change overall model quality evaluated by three metrics: the regression likelihood of all causal relationships on the queried data, Bayesian Information Criterion (BIC)~\cite{burnham2004multimodel}, and p-value. 
The regression likelihood indicates the model goodness of fit, and BIC estimates the complexity of the causal model to ensure better generalization capabilities. The p-value evaluates the significance of improvement between two model updates. The circles \rrv{are} positioned in a two-dimensional space defined by the \rrv{number} of model updates on the $x$-axis and the mean regression likelihood on the $y$-axis. The error bar represents the standard deviation of the regression likelihood. The color of the circles \rrv{encodes} the change of \rrv{the} BIC score \rrv{in comparison to the} previous model. Green circles represent \rrv{the} better generalization capability and red circles \rrv{represent} \rrv{the} worse. The detailed \rrv{values} of these metrics are displayed in a tooltip activated when the user hovers the mouse \rrv{over a} circle. When the performance of the model declines, the user can revert the causal model and causal graph to a previous savepoint by clicking on the circles.

\vspace{-2mm}
\subsection{Causal Comparison}
\name also supports comparing causalities of different sequence subgroups~(\textbf{R7}) in the \textit{causal comparison view}~(Fig.~\ref{fig:ui}(6)). \x{The user can leverage the comparison result to characterize different groups of sequence entities and make customized decisions accordingly. For example, in medical cases, treatment may cause the cure of a disease for one group of patients but not the other. This can be reflected in the difference \rrv{between} the corresponding causal relations.} As mentioned in Section~\ref{sec:causal_verif}, the user can save the final causal analysis result of the queried dataset to the \textit{analysis history view}. In \rrv{this view}, each item shows a general description of the analyzed dataset according to its querying condition. The detailed descriptions, including the user's editing history, statistics on model performance, and the causal graph can be retrieved by expanding the item. 

The user can drag any two items from the \textit{analysis history view} into the \textit{causal comparison view} to compare the causal relations in different subgroups.
\x{We utilize a superimposed adjacency matrix to visualize the occurrence of all causal relations in two groups. The rows of the matrix represent causes and the columns represent \rrv{effects}. The encoding of each cell shows the existence of a causal relation in two subgroups. \rrv{As illustrated} in Fig.~\ref{fig:ui}(6), each cell is divided into an outer region and an inner region, with the background color saturation representing the causal strength of the corresponding causal relation in the first group and the second group, respectively. The encoding of the cell can distinguish a total of five situations of the comparison result (shown in Fig.~\ref{fig:ui}(6a--6e), respectively): (a) the causal relation only exists in the first group, (b) the causal relation only exists in the second group, (c) the causal relation exists in both groups but with different causal strength, (d) the causal relation exists in both groups and has the same causal strength, (e) the causal relation does not exist in both groups. In this case, the user can quickly detect the causal relations that have \rrv{a} significant difference in two groups of sequences.}

\vspace{-1.5mm}
\section{Evaluation}
\begin{figure}[t!]
    \centering
    \includegraphics[width=\linewidth]{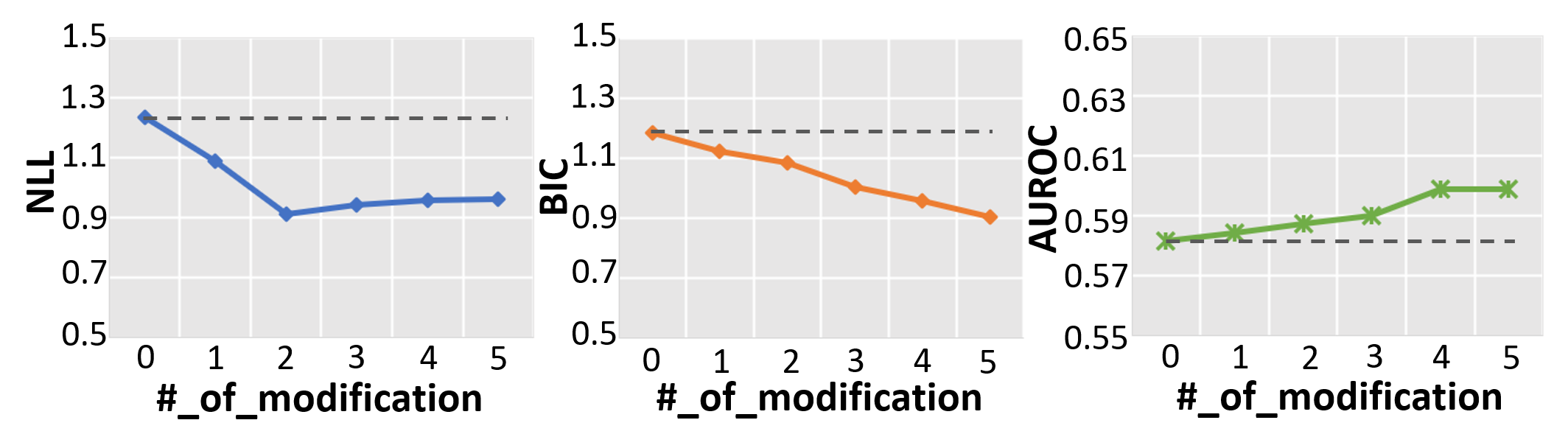}
    \vspace{-8mm}
    \caption{The performance of the user-feedback mechanism under three metrics: negative log-likelihood (blue), Bayesian Information Criterion (orange), and Area Under ROC (green). The dashed lines show the original model performance on the respective metric without user-feedback.}
    \label{fig:per}
    \vspace{-5.5mm}
\end{figure}
We assess the usefulness of \name through two forms of evaluations: a quantitative study showing the effectiveness of the user-feedback mechanism incorporated in the analysis algorithm, and qualitative case studies demonstrating the usefulness of \name system. 
\x{In the quantitative study, we used a public news media dataset, MemeTracker~\cite{Leskovec2009Meme}, which has the ground-truth causality for us to measure the accuracy of the causality analysis result. For the qualitative study, we applied two datasets for distinct applications: a public-access intensive care dataset, MIMIC~\cite{johnson2016mimic}, and a media dataset~\cite{subreddit} that captures users' commenting trajectory on Reddit. These datasets, however, do not contain ground-truth causality. Therefore, we leverage human knowledge to justify the causality analysis results. In this section, we report our study findings and discuss feedback from study participants.}

\vspace{-1.5mm}
\subsection{Performance of User-Feedback Mechanism}
% The data we used in this study is a public news media dataset, MemeTracker~\cite{Leskovec2009Meme}, which contains time stamped phrases and hyperlinks for news articles and blog posts from mainstream media sites. Each sequence records the trace of a \q{meme}(i.e., a representative quote or a phrase) across various websites. 
\x{We employ MemeTracker dataset to evaluate the effectiveness of the user-feedback mechanism, which contains time-stamped phrases and hyperlinks for news articles and blog posts from mainstream media sites. Each sequence records the trace of a meme (i.e., a representative quote or a phrase) across various websites. }
Each event represents an occurrence of a meme on a website, and the website represents the event type. We filter 20 event types of the most active websites and sequence records from August 2008 to September 2008 to train the causality analysis model, and generate causal relations among websites to imply the spreading patterns of memes. The ground-truth causality \rrv{was} provided by whether one website contains the hyperlink linking to another site \cite{achab2017uncovering,xiao2019learning}. 
% \x{The performance of the model starts to stabilize after five iterations of updates, therefore, we display 
% We update the model through five iterations of feedback, as the ground truth causalities of the MemeTracker dataset is relatively simple and the performance of the model starts to stabilize after .}
% The causal relations output from the model were modified through five iterations. 
In each iteration, we stimulated the user feedback by confirming one causal relation in the ground-truth set to update the model. \x{According to the \textit{model diagnostics panel}, the performance of the model starts to stabilize after five iterations. We report the performance changes in the first five iterations from two aspects: the goodness-of-fit and model accuracy. }
% the capability of the user-feedback mechanism in improving the causal analysis model }In the following, we report the capability of the user-feedback mechanism in improving the causal analysis model from two aspects: the goodness-of-fit and model accuracy.

\textbf{Goodness-of-fit.} 
We utilize the negative log-likelihood (NLL) and the \rrv{BIC score} to examine the effect of the user-feedback mechanism on the goodness-of-fit of the causality analysis model. In particular, a smaller NLL value reflects a better fit of the given dataset, and a smaller BIC score indicates lower model complexity and better robustness. As shown in Fig.~\ref{fig:per}, the NLL value significantly decreased in the first two iterations and slightly bounced back afterward. The BIC score, however, keeps declining across all five iterations. This result indicates that the user-feedback mechanism generally improves the goodness-of-fit of the causality analysis model.

\textbf{Accuracy.}
We utilize the Area Under ROC (AUROC) to evaluate the effect of the user-feedback mechanism on the improvement of accuracy. Note that the ground-truth causality provided by human modification is \rrv{excluded} when calculating AUROC in each iteration so that the value is only influenced by \rrv{the change of causalities from the model}. Higher AUROC indicates a better accuracy of the causality analysis result against the ground-truth causality. As shown in Fig.~\ref{fig:per}, the AUROC value gradually increases as the user provides valid corrections to the causality analysis result. This observation indicates that the performance of the model in terms of inferring accurate causality can be improved by the user-feedback mechanism if the feedback provided by the user is correct.

\vspace{-1.6mm}
\subsection{\x{Case Studies}}
\x{We demonstrate the usability of \name through two case studies in different application scenarios using electronic health records and social media interactions, respectively. }

\begin{figure}[t!]
    \centering
    \includegraphics[width=\linewidth]{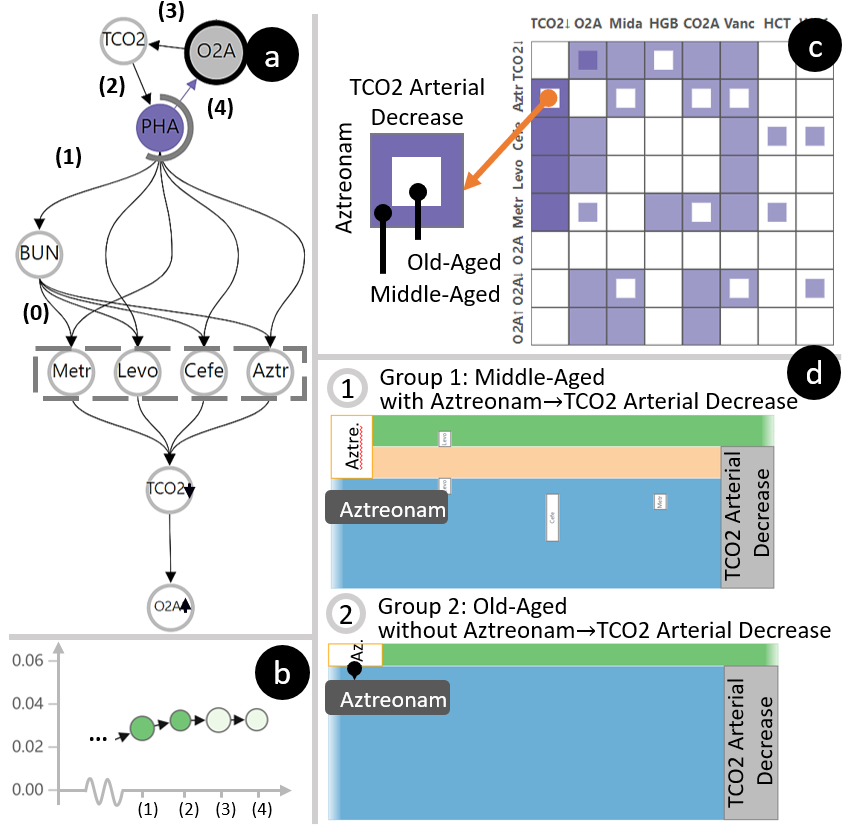}
    \vspace{-8mm}
    \caption{The causality of pneumonia symptoms and treatments in MIMIC dataset. (a-b) Generating \rrv{a} causality graph for middle-aged pneumonia patients. (c-d) Comparing \rrv{the} causalities of middle-aged and old-aged pneumonia patients with evidence from raw data.}
    \vspace{-6mm}
    \label{fig:case1}
\end{figure}

\vspace{-2mm}
\subsubsection{Causality in Electronic Health Records}
\x{This case study employs the MIMIC dataset, which contains electronic health records of over 46,000 patients with various diseases.} We invited two pulmonologists (\textbf{E1, E2}) with more than 8 years of clinical experience to participate in our case study. In particular, the medical experts were also involved in determining the design requirements of \name discussed in Section~\ref{sec:overview}. Prior to the study, we asked the pulmonologists to identify a list of key event types that might be causally related for analysis, which includes 120 events under the category of laboratory tests and medical treatments. \x{Since all variables in our causality analysis model are discrete events, we preprocessed continuous laboratory tests by filtering out the normal records and discretizing the abnormal value by \rrv{whether} it increases or decreases compared to the previous record. Note that the increase and decrease of \rrv{values} only represent the occurrence of discrete events instead of directions of the causal relation. We encode three situations of the laboratory tests by varying their labels: the abnormal values with no previous record are labeled by the name of the lab test event, and the value increase and decrease are labeled with an ascending or descending arrow in the end. } 

\x{The study started with a 20-minute introduction \rrv{of} the system and visualization design. Then, the doctors took an hour exploring our system \rrv{and} \rrv{demonstrated} their findings in a think-aloud manner. Finally, we conducted a 30-minute post-study interview collecting \rrv{the} doctors' subjective comments on the system's usability.} 
% The doctors were then asked to explore the system and demonstrate their findings during the study in a think-aloud manner. \x{The study session lasted for approximately 1.5 hour, including one-hour evaluation of the system and a half-hour interview collecting the doctors' subjective comments on the system's usability.} 
In the following, we report two representative insights and discuss feedback from \rrv{the} experts. 

\textbf{Causality of pneumonia symptoms.}
\x{The doctors queried a group of 127 middle-aged patients aging from 50 to 60 who were diagnosed with pneumonia. The retrieved dataset contains 93 types of events. They started by adding \textit{O2 arterial increase} as an outcome event to explore its causes, which is an important sign of recovery for pneumonia patients. After several iterations of confirming causalities and model updates, the doctors noticed that \textit{abnormal BUN} value was identified as the cause of taking four treatments for improving renal functionality (Fig.~\ref{fig:case1}(a-0)). This is in line with their domain knowledge as the abnormality in \textit{BUN} indicates renal damage. In addition, the system suggested \rrv{that} \textit{abnormal arterial pH} was a potential cause for \textit{BUN} anomaly (Fig.~\ref{fig:case1}(a-1)). After inspecting the \textit{causal sequence view}, where half of the patients with \textit{abnormal arterial pH} are also abnormal in the test of \textit{BUN}, the doctors confirmed this causal relation, and the regression likelihood of the model was improved (Fig.~\ref{fig:case1}(b)). The doctors further examined the cause of \textit{abnormal arterial pH} and found a causality circle among three laboratory indices after three iterations of update: \textit{O2}, \textit{TCO2}, and \textit{pH} values in \rrv{the} artery. The doctors confirmed the causality circle and explained: }\q{For patients with pneumonia, the value of oxygen, the value of carbon dioxide, and the value of \textit{pH} in blood always affect each other. Because of this cyclical causality, the conditions of patients will keep getting worse [if not intervened].} At this point, the causal chain for the outcome event \textit{O2 arterial increase} is complete, and the doctors saved the final causality to \rrv{the} \textit{analysis history view}.

\textbf{\x{Effect} of antibiotic medicines in different cohorts.}
\x{After analyzing \rrv{the} causalities of middle-aged patients, the doctors further queried a group of old-aged patients aging from 80 to 90 for comparison, as they anticipated that the effect of antibiotic therapy \rrv{might} differ between age groups. The queried dataset contains 174 sequences and 79 event types. As shown in Fig.~\ref{fig:case1}(c), both groups have causal relations that link toward the decrease of \textit{arterial TCO2}, which is an important indicator for the improvement of the patient's condition. However, the doctors found that the use of \textit{Aztreonam} was effective in the middle-aged cohort, \rrv{whereas} the old-aged cohort did not have such causal relation.
This can also be observed from the \textit{causal sequence view}, where \textit{Aztreonam} seemed effective to half of the middle-aged cohort (Fig.~\ref{fig:case1}(d-1)) but none of the old-aged cohort (Fig.~\ref{fig:case1}(d-2)). In addition, the middle-aged group seemed to have multiple choices of antibiotic treatments. As shown in Fig.~\ref{fig:case1}(d-1), a large group of patients with \rrv{a} decrease in \textit{arterial TCO2} had not taken \textit{Aztreonam}.} \textbf{E2} found this reasonable, as he said: \q{Elderly patients are normally weak and suffer from many other complications. It \rrv{needs} to be particularly careful to apply antibiotics medicine to them.}

\x{\textbf{Expert feedback.}} Both experts felt that the query view is very useful in medical \rrv{scenarios} for filtering a cohort with similar conditions (\textbf{R1}). They suggested that more detailed filters \rrv{could} be added, such as ranges of some key laboratory tests. The doctors also felt the design of \rrv{the} \textit{causal graph view} and the \textit{causal sequence view} \rrv{could} help them explore and verify causalities in medical events efficiently (\textbf{R2, R4}). As \textbf{E1} commented:\q{This system can help us discover potentially causal related medical events or spurious causal relations, and allow us to verify the relations in the original data efficiently.} \x{\textbf{E2} found that personally confirming the causal relations enhances his confidence on the causality analysis result, especially with the performance of the model displayed in the \textit{model diagnostics view} (\textbf{R5, R6})}.

\x{The experts also commented on the visualization of our system.} \textbf{E1} applauded the layout scheme of the causal graph, as he said: \q{The causal chains are easy to trace and the cyclical structure is properly emphasized.} He also found the design of \rrv{the} \textit{causal sequence view} useful for interpreting and verifying the causal relations (\textbf{R3}): \q{In this view, we can easily infer the causes and effects from original data. It is a good way to interpret and verify the causal relations.} In terms of comparing causal relations (\textbf{R7}), both experts agreed that it is convenient to retrieve previous analysis results from the \textit{analysis history view}. However, the experts felt that the matrix-based design is a bit overwhelming, and \textbf{E1} suggested that: \q{Using text descriptions to illustrate the differences may be easier to understand.}

\subsubsection{\x{Causality of Subreddit Interactions}}
\x{We also applied our system in analyzing the sequences of user comments on different subreddits from 2016 to 2017. Each subreddit represents a sub-community of the Reddit community with a particular area of interest. We extracted each user's commenting trajectory on various subreddits as an event sequence, where each event is a subreddit with \rrv{a} timestamp. We invited a Reddit user who is familiar with the 
characteristics of different sub-communities to analyze how the traffic of different subreddits \rrv{is} mutually affected \rrv{by} their causalities. The study lasted 45 minutes, including \rrv{a} 20-minute system introduction and \rrv{a} 25-minute causality analysis using our system. The participant showed his interest in analyzing how Reddit users \rrv{were} attracted to \textit{The\_Donald}, which was created in support of U.S. President Donald Trump and became particularly popular during the presidential campaign in 2016. Therefore, he queried Reddit users who had commented under \textit{The\_Donald} by adding it as the key event. The system retrieved 204 sequences with 165 event types for analysis. }

\begin{figure}[t!]
    \centering
    \includegraphics[width=0.93\linewidth]{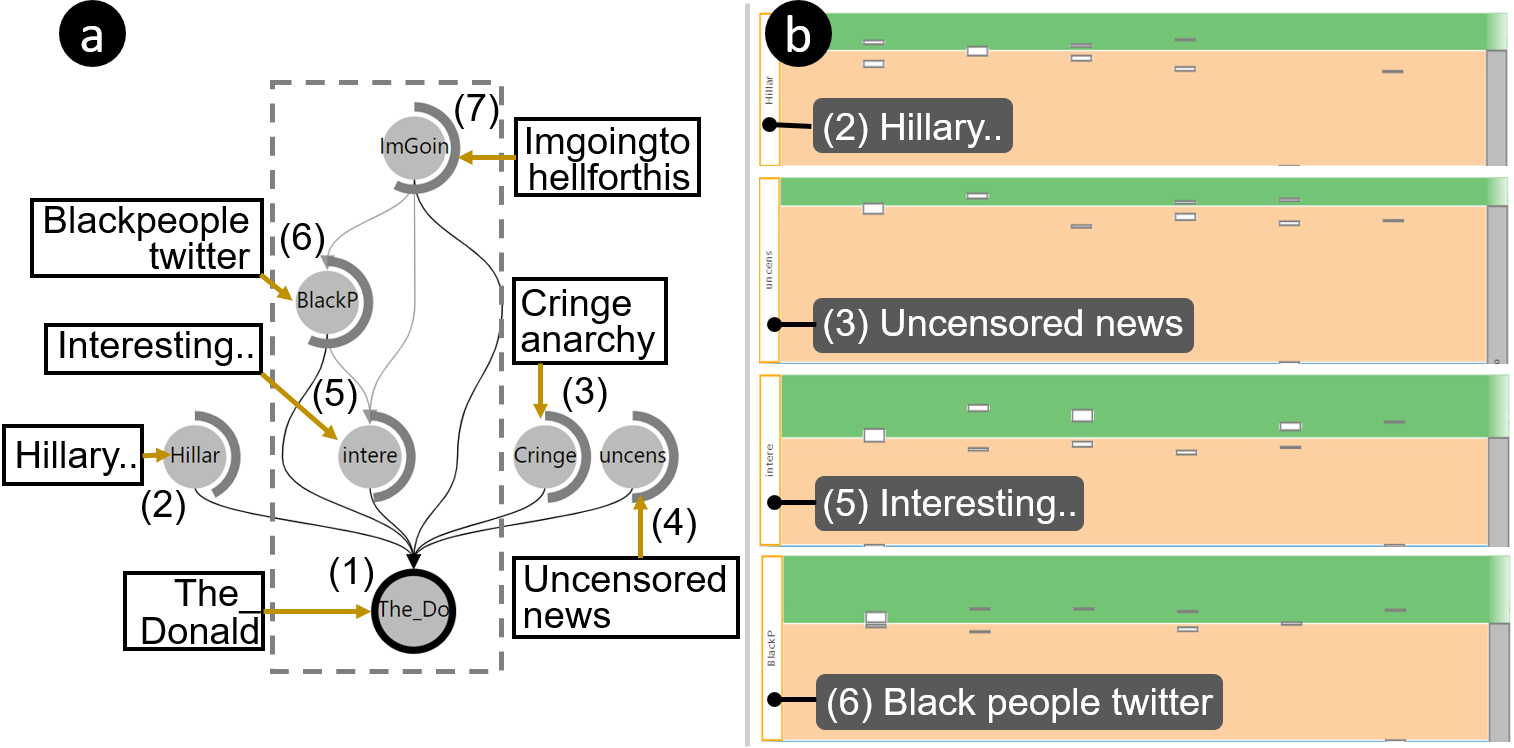}
    \vspace{-4.3mm}
    \caption{The causality of subreddit interactions showing how users from other subreddits are attracted to \textit{The\_Donald}.}
    \vspace{-5.6mm}
    \label{fig:case3}
\end{figure}

\x{He initiated the exploration of causalities by adding \textit{The\_Donald} as the outcome event~(Fig.~\ref{fig:case3}(a-1)). To eliminate noisy causal relations, he raised the threshold of event coverage to make sure that each causal relation exists in at least 30\% of the queried sequences. After three iterations of \rrv{the} model update, new events were no longer added to the causality graph. Fig.~\ref{fig:case3} shows the final state of the causality graph. The participant found the traffic of \textit{The\_Donald} mainly came from three sources: \textit{Hillary...} (Fig.~\ref{fig:case3}(a-2)), a subreddit of negative comments from Hillary Clinton's opponents, who was also participated in the presidential election in 2016; \textit{CringeAnarachy} and \textit{UncensoredNews} (Fig.~\ref{fig:case3}(a-(3,4)), two subreddits of mostly \rrv{politically} related news; and several popular subreddits of anecdotes, including \textit{Interesting...}, \textit{BlackPeopleTwitter}, and \textit{ImGoingHellForThis} (Fig.~\ref{fig:case3}a-(5,6,7)). He then turned to the \textit{causal sequence view} to check the validity of causal relations. Although all subreddits in the graph can directly cause \textit{The\_Donald} according to the causality analysis result, the causal patterns in raw event sequences \rrv{indicate} the difference of these causal relations. In particular, the participant found that subscribers of \rrv{politically} related subreddits had \rrv{a} smaller group of sequences with A\textrightarrow ? pattern comparing with anecdotes subscribers. Examples are illustrated in Fig.~\ref{fig:case3}(b): subsequences correspond to causes \textit{Hillary...} and \textit{UncensoredNews} have narrower green edges compared to the causes \textit{Interesting...} and \textit{BlackPeopleTwitter}. This indicates that \textit{Hillary...} and \rrv{politically} related news are more likely to be valid causes of the \textit{The\_Donald}'s popularity.}

% \begin{figure}[t!]
%     \centering
%     \includegraphics[width=\linewidth]{figures/case1.png}
%     \vspace{-8mm}
%     \caption{The causal relations among symptoms and treatments of pneumonia patients discovered by the pulmonologists. Two major causality patterns were uncovered from the causal graph (a): the abnormal \textit{O2}, \textit{TCO2}, and \textit{pH} value in arterial are mutually effected, forming a vicious causality circle~(a--2,3,4); four medical treatments are applied to cure the disease~(a--0). The causal patterns were verified from the raw event sequences (b), and the performance of the causal model was gradually improved as the doctors confirmed the causal relations (c).}
%     \vspace{-5mm}
%     \label{fig:case1}
% \end{figure}

\vspace{-2mm}
\section{Discussion}
\label{sec:dis}
This section includes \rrv{a} discussion on the generalizability of our system, the scalability of our causality analysis algorithm, the limitations of the current study, and promising future directions.

\textbf{Generalizability.} 
Although \x{the design requirements} of \name \x{\rrv{were} collected from} \rrv{the} medical domain, the causality analysis algorithm and visualization \x{\rrv{were} designed for general event sequence analysis} and can be easily generalized. For sequences of discrete events in continuous time (e.g., web clickstreams, social media actions, etc.) where events are not observed in fixed time lags, our causality analysis algorithm can be directly utilized, whereas for fixed time-lagged sequences (e.g., text streams, discretized time-series), the sampling function $\kappa(t)$ as mentioned in Section~\ref{sec:hawkes} \rrv{needs} to be replaced with Poisson sampling to better fit events in discrete time. The visualization design of our system is not tailored to \rrv{a} specific application domain and can be directly applied to any event sequence dataset.

\begin{table}[]
\centering
\resizebox{0.70\linewidth}{!}{
\begin{tabular}{|c|ccc|}
\hline
                                                                               & \multicolumn{3}{c|}{\# of occurrences for   all events ($n$)} \\ \cline{2-4} 
\multirow{-2}{*}{\begin{tabular}[c]{@{}c@{}}\# of event \\ types ($V$)\end{tabular}} & 5334                     & 10668                     & 16002                    \\ \hline
31                                                                             & 1.82$\pm$0.02                     & 3.75$\pm$0.02                    & 5.54$\pm$0.02                     \\
62                                                                             & 1.88$\pm$0.02                     & 3.75$\pm$0.02                      & 5.49$\pm$0.02                     \\
93                                                                             & 1.83$\pm$0.02                     & 3.7$\pm$0.03                       & 5.56$\pm$0.03                     \\ \hline
\end{tabular}}
\vspace{0.1mm}
\caption{The running time of the causal modeling algorithm under different \rrv{numbers} of event types and event occurrences.}
\label{table:running_time}
\vspace{-8mm}
\end{table}

\textbf{Scalability.}
\x{We \rrv{tested} the scalability of our causal modeling algorithm with nine synthetic datasets of different \rrv{numbers} of event types and event occurrences. The synthetic datasets were generated \rrv{by} modifying a MIMIC case study dataset. We recorded the running \rrv{times} on a Linux server with an Intel Xeon CPU (GD6148 2.4 GHz/20-cores) and 192 GB RAM. As illustrated in Table~\ref{table:running_time}, the running time increases with the number of event occurrences. However, it is independent of the number of event types, benefiting from the parallel computation of events described in Section~\ref{sec:user-feedback}. 
% In our MIMIC case study, It took $1.83 \pm 0.02$ seconds to generate the initial causalities and $1.88    \pm 0.02$ seconds in each iteration to update the result.
Although the periodic delay was not noticed as a problem by our expert users in the case study, the system may become difficult to interact \rrv{in} real-time as the number of event occurrences grows (i.e., the length or the number of sequences becomes larger). This requires further research for more efficient tuning of the causality analysis algorithm.}

\x{The scalability issue also exists in visualizing and exploring the causal relations (\textbf{R2}). \rrv{Although} we mitigate the problem by introducing the layout algorithm and user-driven causality exploration procedure described in Section~\ref{sec:causal_dis}, the growing number of event types can increase the complexity of the causal graph displayed in the \textit{causal model view}, making it difficult for the user to visually explore and interact with. Our current design cannot fully support \rrv{the} analysis of event sequence data with very high dimensionality. A more scalable visualization and efficient interaction mechanism for high-dimensional causal \rrv{graphs} are required in the future research.}

\textbf{Limitations and future work.}
% In addition to the scalability issue of the causality analysis algorithm, 
\x{In addition to the scalability issue,}
we also recognize several other limitations of our work that point toward promising future directions. First, our system currently supports exploring causalities from effects to search for the causes. However, our medical experts suggested that it is also meaningful to \x{investigate potential effects from causes for prognostic analysis.}
% use causes as starting points for discovering the potential effects, as what they did in prognostic analysis. 
Allowing bi-directional exploration of the causal relations would increase the exploring space, which requires more efficient interaction methods to be studied in the future. 
% Secondly, although our system is generalizable to any types of event sequence data, the case study was only conducted in the medical domain. 
\x{In addition, our causality analysis algorithm is not capable of discovering combined causes. For example, in the case where two treatments together cause the cure of a symptom, our system identifies each treatment as an individual cause. Although there are some causality analysis algorithms that are capable of mining combined causes~\cite{MA2016104,azizi2014learning}, they mainly focus on analyzing non-temporal data and cannot be directly applied to temporal event sequences. We still need to explore more advanced causality analysis \rrv{algorithms} and design corresponding visualizations to support accurate and efficient analysis of combined causes for event sequence data.} 
% \x{Thirdly, although our system is designed to be generalizable to any types of event sequence data, our requirement analysis is mainly based on the analytical tasks of medical domain.} In the future, we need long-term case studies to examine our system in different application scenarios and advance our understanding of the design requirements in more causality analysis tasks.
\vspace{-2mm}
\section{Conclusion}
In this paper, we have presented \name, an interactive visual analytics system for analyzing causalities in \rrv{the} event sequence dataset. The system employs a Granger causality analysis algorithm based on Hawkes processes with a user-feedback mechanism to leverage human knowledge in revising the causality analysis model. Analysts can utilize the system to discover causal relations of events, investigate complex causalities with efficiency through the causal exploration, verification, and comparison. Our quantitative study has demonstrated that the goodness-of-fit and accuracy of the model can be iteratively improved with our user-feedback mechanism. The case studies have shown the capabilities of our system in helping experts extract interesting insights into potential causal related events and discover useful causal patterns. 

\vspace{-2mm}
\acknowledgments{
Nan Cao is the corresponding author. We would like to thank all the study participants, domain experts, and reviewers for their valuable comments. This work was supported in part by the Fundamental Research Funds for the Central Universities (Grant No.22120190216).}

\bibliographystyle{abbrv-doi}

\balance

\end{document}